\documentclass[runningheads]{llncs}
\usepackage{graphicx}
\usepackage{amsmath,amssymb} 
\usepackage{multirow}
\usepackage{hyperref}
\makeatletter
\newcommand{\printfnsymbol}[1]{%
  \textsuperscript{\@fnsymbol{#1}}%
}
\makeatother

\begin{document}
%




\title{SketchyScene: Richly-Annotated Scene Sketches} 

\author{Changqing Zou\thanks{equal contribution}$^1$  \quad Qian Yu\printfnsymbol{1}$^{2}$  \quad Ruofei Du$^1$ \quad Haoran Mo$^3$ 
\\ Yi-Zhe Song$^2$ \quad Tao Xiang$^2$ \quad Chengying Gao$^3$ \quad \\ Baoquan Chen\thanks{corresponding author: baoquan@sdu.edu.cn}$^4$ \quad  Hao Zhang$^5$ \\
University of Maryland, College Park, United States$^1$  \\
Queen Mary University of London, United Kingdom$^2$ \\
Sun Yat-sen University, China$^3$ \quad Shandong University, China$^4$\\
Simon Fraser University, Canada$^5$
}
\authorrunning{C. Zou et al.}
\institute{}

%

\maketitle
\vspace{-1.5cm}

\begin{abstract}

\begin{figure}[h!]
    \centering
    \includegraphics[width=0.95\textwidth]{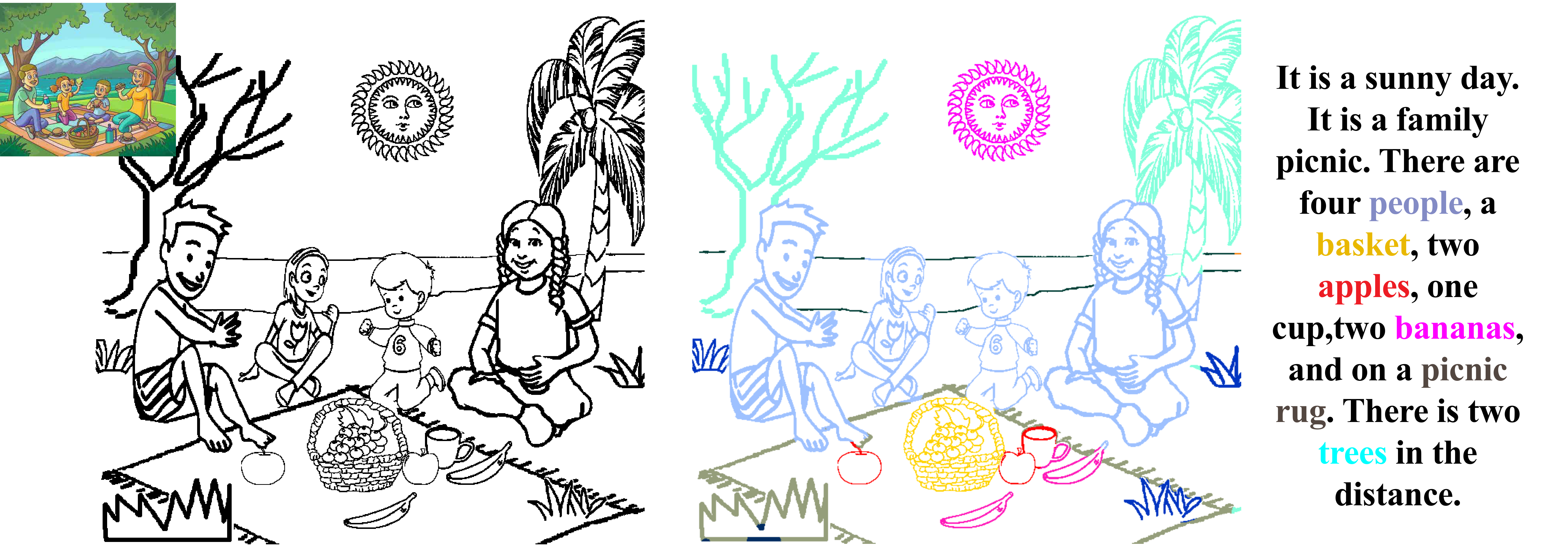}
    \caption{A scene sketch from our dataset {\sc SketchyScene\/} that is user-generated based on the reference image shown, a segmentation result (middle) obtained by a method trained on {\sc SketchyScene\/}, and a typical application: sketch captioning.}           
    \label{fig:skethGT}
 \end{figure}
 
We contribute the first large-scale dataset of {\em scene sketches\/}, {\sc SketchyScene}, with the goal of advancing research on sketch understanding at both the object and scene level. The dataset is created through a novel and carefully designed {\em crowdsourcing\/} pipeline, enabling users to efficiently generate large quantities of realistic and diverse scene sketches. {\sc SketchyScene} contains more than 29,000 scene-level sketches, 7,000+ pairs of scene templates and photos, and 11,000+ object sketches. All objects in the scene sketches have ground-truth semantic and instance masks. The dataset is also highly scalable and extensible, easily allowing augmenting and/or changing scene composition. We demonstrate the potential impact of {\sc SketchyScene} by training new computational models for semantic segmentation of scene sketches and showing how the new dataset enables several applications including image retrieval, sketch colorization, editing, and captioning, etc. {The dataset and code can be found at https://github.com/SketchyScene/SketchyScene.}
\end{abstract}

\section{Introduction} 

In the age of data-driven computing, large-scale datasets have become a driving force for improving and differentiating the performance, robustness, and generality of machine learning algorithms. In recent years, the computer vision community have embraced a number of large and richly annotated datasets for images (e.g., ImageNET~\cite{deng2009imagenet} and Microsoft COCO~\cite{lin2014microsoft}), 3D objects (e.g., ShapeNET~\cite{chang2015shapenet,wu2015shapenet} and PointNET~\cite{qi2017pointnet}), and scene environments (e.g., SUN~\cite{xiao2010sun} and the NYU database~\cite{silberman2012}). Among the various representations of visual forms, hand-drawn sketches occupy a special place since, unlike most others, they come from human creation. Humans are intimately familiar with sketches as an art form and sketching is arguably the most compact, intuitive, and frequently adopted mechanism to visually express and communicate our impression and ideas.

Significant progress has been made on sketch understanding and sketch-based modeling in computer
vision and graphics recently~\cite{eitz2012hdhso,Schneider2014,ZouYL14,WangKL15,dekel2017smart,song2017deep,hu2018sketch,xu2018sketchmate,song2018learning,LiHZTLZTF16,ZhangCLGT18}. Several large-scale sketch datasets~\cite{eitz2012hdhso,sketchy2016,ha2017neural} have also been constructed and utilized along the way.
Nevertheless, these datasets have all been formed by {\em object} sketches and the sketch analysis and processing tasks have mostly been at the stroke or object level. 
Extending both to the {\em scene} level is a natural progression towards a deeper and richer reasoning about sketched visual forms. 
The ensuing analysis and data synthesis problems become more challenging since a sketched scene may contain numerous objects interacting in a complex manner. While scene understanding is one of the hallmark tasks of computer vision, the problem of understanding scene sketches have not been well studied.

\begin{figure}[!t]
   \centering
      \includegraphics[width=\textwidth]{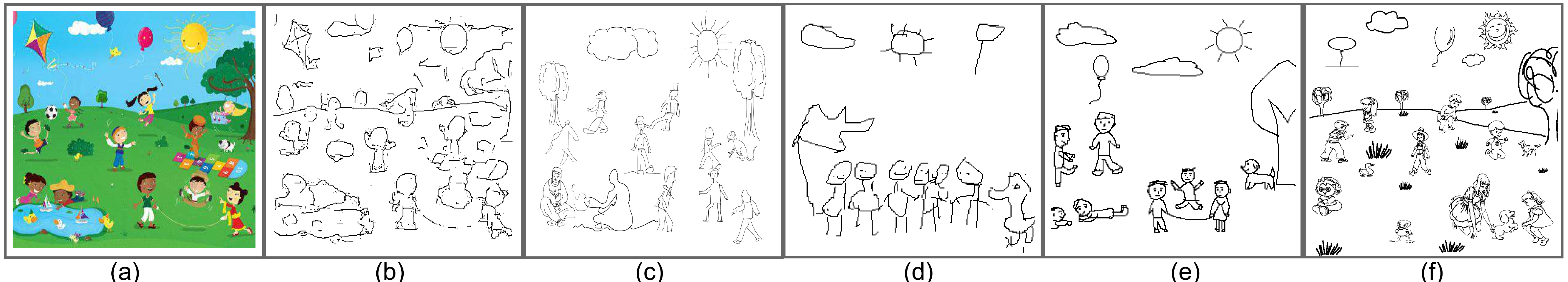}
      \caption{(a) reference image; (b) response of edge detector; (c) synthesized scene using object sketches from Sketchy and TU-Berlin (using the same pipeline as (f)); (d) non-artist's drawing with the hint of a short description; (e) artist's drawing with the hint of reference image; (f) synthesized scene using our system. Processes of (c)-(f) take 6, 8, 18, and 5 minutes, respectively.}
      \label{fig:drawingStrategies}
  \end{figure}

In this paper, we introduce the first large-scale dataset of {\em scene sketches\/}, which we refer to as {\sc SketchyScene}, to facilitate sketch understanding and processing at both the object and scene level. Obviously, converting images to edge maps~\cite{XieT15} does not work since the results are characteristically different from hand-drawn sketches. Automatically composing existing object sketches based on predefined layout templates and fitting the object sketches into stock photos are both challenging problems that are unlikely to yield a large quantity of realistic outcomes (see Fig. \ref{fig:drawingStrategies}(b)). In our work, we resort to {\em crowdsourcing\/} and design a novel and intuitive interface to reduce burden placed on the users and improve their productivity. Instead of asking the users to draw entire scene sketches from scratch, which can be tedious and intimidating, we provide object sketches so that the scene sketches can be created via simple interactive operations such as drag-n-dropping and scaling the object sketches. To ensure diversity and realism of the scene sketches, we provide reference images to guide/inspire the users during their sketch generation. With a user-friendly interface, participants can create high-quality scene sketches efficiently. On the other hand, the scene sketches synthesized this way are by and large {\em sketchy sketches}~\cite{eitz2012hdhso,sketchy2016} and they do not quite resemble ones produced by 
professional artists.


{\sc SketchyScene} contains both object- and scene-level data, accompanied with rich annotations. In total, the dataset has more than {\em 29,000 scene sketches\/} and more than {\em 11,000 object sketches\/} belonging to 45 common categories. In addition, more than {\em 7,000 pairs of scene sketch templates and reference photos\/} and more than {\em 200,000 labeled instances\/} are provided. Note that, {\em all\/} objects in the scene sketches have ground-truth semantic and instance masks. More importantly, {\sc SketchyScene} is {\em flexible\/} and {\em extensible} due to its object-oriented synthesis mechanism. Object sketches in a sketch scene template can be switched in/out using available instances in {\sc SketchyScene} to enrich the dataset. 


We demonstrate the potential impact of {\sc SketchyScene} through experiments. Foremost, the dataset provides a springboard to investigate an assortment of problems related to scene sketches (a quick Google Image Search on ``scene sketches" returns millions of results). In our work, we investigate semantic segmentation of scene sketches for the first time. To this end, we evaluated the advanced natural image segmentation model, DeepLab-v2~\cite{CP2016Deeplab}, exploring the effect of different factors and providing informative insights. We also demonstrate several applications enabled by the new dataset, including sketch-based scene image retrieval, sketch colorization, editing, and captioning.



\section{Related work}

\subsection{Large-scale sketch datasets}
\label{subsec:dataset}
There has been a surge of large-scale sketch datasets in recent years, mainly driven by applications such as sketch recognition/synthesis~\cite{eitz2012hdhso,ha2017neural} and SBIR \cite{yu2016sketch,sketchy2016}. Yet the field remains relatively under-developed with existing datasets mainly facilitating object-level analysis of sketches. This is a direct result of the non-ubiquitous nature of human sketches data -- they have to be carefully crowd-sourced other than automatically crawled for free (as for photos). 

{TU-Berlin}~\cite{eitz2012hdhso} is the first such large-scale crowd-sourced sketch dataset which was primarily designed for sketch recognition. It consist of 20,000 sketches spanning over 250 categories. The more recent {QuickDraw}\cite{ha2017neural} dataset is much larger, with 50 million sketches across 345 categories. Albeit being large enough to facilitate stroke-level analysis\cite{chen2017sketch}, sketches sourced in these datasets were produced by sketching towards a semantic concept (e.g., ``cat", ``house"), without a reference photo or mental recollection of natural scene/objects. This greatly limits the level of visual detail and variations depicted, therefore making them unfitting for fine-grained matching and scene-level parsing. For example, faces are almost all in their frontal view, and depicted as a smiley in QuickDraw.

The concurrent work of \cite{yu2016sketch} and \cite{sketchy2016} progressed the field further by collecting object instance sketches for FG-SBIR. QMUL database \cite{yu2016sketch} consists of 716 sketch-photo pairs across two object categories (shoe and chair), with reference photos crawled from on-line shopping websites. Sketchy \cite{sketchy2016} contains 75,471 sketches and 12,500 corresponding photos across a much wider selection of categories (125 in total). Object instance sketches are produced by asking crowdsourcers to depict their mental recollection of a reference photo. In comparison with concept sketches~\cite{eitz2012hdhso,ha2017neural}, they by and large exhibit more object details and have matching poses with the reference photos. However, a common drawback for both, for the purpose this project, lies with their limited pose selection and object configurations. QMUL sketches exhibit only one object pose (side view) under a single object configuration. Scene sketches albeit exhibits more object poses and configurations, are still restricted since their reference photos mainly consists of single objects centered on relatively plain backgrounds (thus depicts no object interactions). This drawback essentially renders them both unsuitable for our task of scene sketch parsing, where complex mutual object interactions dictate high degree of object pose and configuration variations, as well as subtle details. For example, within a picnic scene depicted in Figure~\ref{fig:skethGT}, people appear in different poses and configurations with subtle eye contacts among each other. Fig. \ref{fig:drawingStrategies}(c) shows a composition result using sketches from Sketchy and TU-Berlin. 

{\sc SketchyScene} is the first large-scale dataset specifically designed for scene-level sketch understanding. It differs from all aforementioned datasets in that it goes beyond single object sketch understanding to tackle scene sketch, and purposefully includes an assorted selection of object sketches with diverse poses, configurations and object details to accommodate the complex scene-level object interactions. Although the existing dataset Abstract Scenes \cite{ZitnickP13} serves a similar motivation for understanding high-level semantic information in visual data, they focus on abstract scenes composed using clip arts, which include much more visual cues such as color and texture. In addition, their scenes are restricted in describing interactions between two characters and a handful of objects, while the scene contents and mutual object interactions in {\sc SketchyScene} are a lot more diverse.

\subsection{Sketch understanding}
Sketch recognition is perhaps the most studied problem in sketch understanding. Since the release of TU-Berlin dataset~\cite{eitz2012hdhso}, many works have been proposed and recognition performance had long passed human-level \cite{YuYLSXH17}. Existing algorithms can be roughly classified into two categories: 1) those using hand-crafted features~\cite{eitz2012hdhso,Schneider2014}, and 2) those learning deep feature representation~\cite{YuYLSXH17,ha2017neural}, where the latter generally outperforms the former by a clear margin. Other stream of work had delved into parsing object-level sketches into their semantic parts. \cite{SunWZZ12} proposes an entropy descent stroke merging algorithm for both part-level and object-level sketch segmentation. Huang et al. \cite{HuangFL14} leverages a repository of 3D template models composed of semantically labeled components to derive part-level structures. Schneider and Tuytelaars~\cite{Schneider2016} performs sketch segmentation by looking at salient geometrical features (such as T junctions and X junctions) under a CRF framework. Instead of studying single object recognition or part-level sketch segmentation, this work conducts exploratory study for scene-level parsing of sketches, by proposing the first large-scale scene sketch dataset.

\subsection{Scene sketch based applications}
While no prior work aimed at parsing sketches at scene-level, some interesting applications had been proposed that utilize scene sketches as input. Sketch2Photo~\cite{ChenCTSH09} is a system which combines sketching and photo montage for realistic image synthesis, where Sketch2Cartoon~\cite{sketch2cartoon} is a similar system that works on cartoon images. Similarly, assuming objects have been segmented in a sketchy scene, Xu et al. \cite{XuCFSH13} proposed a system named sketch2scene which automatically generates 3D scenes by aligning retrieved 3D shapes to segmented objects in 2D sketch scenes. Sketch2Tag~\cite{SunWZZ12a} is a SBIR system where scene items are automatically recognized and used as a text query to improve retrieval performance. Without exception, all aforementioned applications involve manual tagging and/or segmentation of sense sketches. In this work, we provide means of automatic segmentation of scene sketches, and demonstrate the potential of the proposed dataset by proposing a few novel applications. 

\section{{\sc SketchyScene} Dataset}
\label{sec:datasets}

A scene sketch dataset should reflect the scenes with sufficient diversity, in terms of their configurations, object interactions and subtle appearance details, where sketch should also contain multiple objects of different categories. 
Besides, the volume of a dataset is important, especially in the context of deep learning. However, as previously discussed, building such dataset based on existing datasets is infeasible, while collecting data from humans can be expensive and time-consuming, therefore an efficient and effective data collection pipeline is required. 




The easiest solution is to ask people to draw a scene directly with provided objects or scene labels as hints (i.e., the strategy used in \cite{eitz2012hdhso}). Unfortunately, this method has proven to be infeasible in our case: (1) Most people are not trained artists. As a result, they struggled to draw complex objects present in a scene, especially when they are in different poses and object configurations (see Fig.~\ref{fig:drawingStrategies}(d)); (2) although different people have different drawing styles, people tend to draw a specific scene layout. For example, given the hint ``\textit{several people are playing on the ground, sun, tree, cloud, balloon and dog}'', people always draw the objects along a horizontal line. That makes the collected scene sketches monotonous in layout and sparse in visual feature. (3) Importantly, this solution is unscalable  -- it takes average person around 8 minutes to finish a  scene sketch of reasonable quality, where costing ~18 minutes for a professional (see Fig.~\ref{fig:drawingStrategies}(e)). This will prohibit us from collecting a large-scale dataset.  

A new data collection strategy is thus devised which is to synthesize a sketchy scene by composing provided object components under the guidance of a reference image. The whole process includes three steps.


\noindent\textbf{Step1: Data Preparation.}
We selected 45 categories for our dataset, including objects and stuff classes. Specifically, we first considered several common scenes (e.g., garden, farm, dinning room, and park) and extracted 100 objects/stuff classes from them as raw candidates. Then we defined three super-classes, i.e. Weather, Object, and Field (Environment), and assigned the candidates into each super-class. Finally, we selected 45 from them by considering their combinations and commonness in real life. 


 \begin{figure}[!t]
  \centering
     \includegraphics[width=0.95\textwidth]{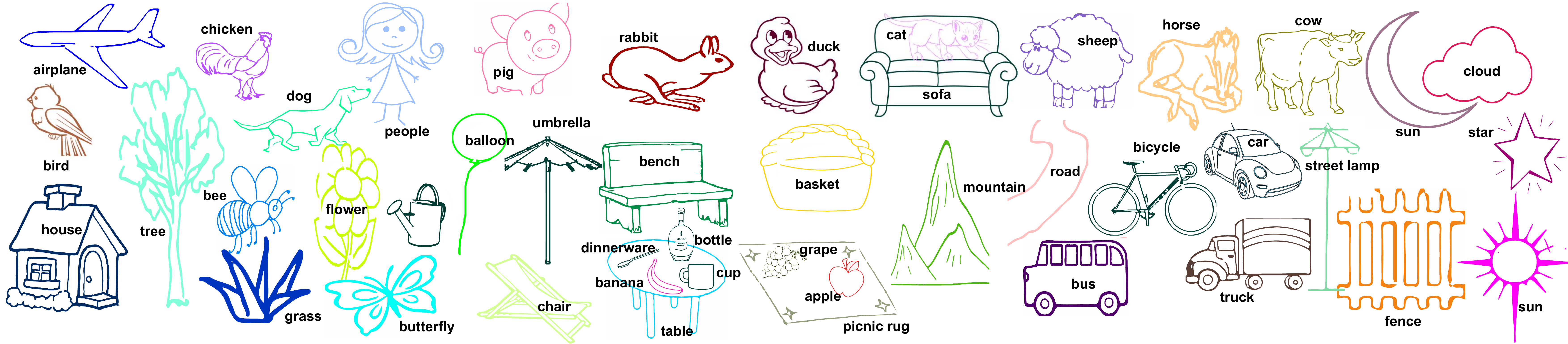}
     \caption{Representative object sketches of {\sc SketchyScene\/}.}
     \label{fig:gallary}
 \end{figure}

Instead of asking workers to draw each object, we provided them with plenty of object sketches (each object candidate is also refer to a ``component") as candidates. In order to have enough variations in the object appearance in terms of pose and appearance, we searched and downloaded around 1,500 components for each category. Then we employed 5 experienced workers to manually sort out the sketches containing single component or cutout individual component from sketches having multiple components. For some categories with  few searched components ($<$20) like ``umbrella", components were augmented through manual drawing. We totally collected {11,316} components for all 44 categories (excluding ``road", which are all hand-drawing, and ``others"). These components of each category are split into three sets: training (5468), validation (2362), and test (3486). Representative components of the 45 categories are shown in Figure~\ref{fig:gallary}. 

In order to guarantee the diversity of the scene layout in our dataset, we additionally collected a set of cartoon photos as reference images. Through sampling the class label(s) from each of our predefined super-classes, e.g., sun (Weather), rabbit (Object), mountain (Environment), we generated 1,800 query items \footnote{We add another label ``cartoon" to each query in order to retrieve cartoon photos}. And around 300 cartoon photos were retrieved for each query items. After removing the repeated ones manually, there are 7,264 reference images (4730 images are unique). These reference images are also split into three sets for training (5,616), validation (535), and test (1,113).    

\begin{figure}[!t]
  \centering
     \includegraphics[width=0.95\textwidth]{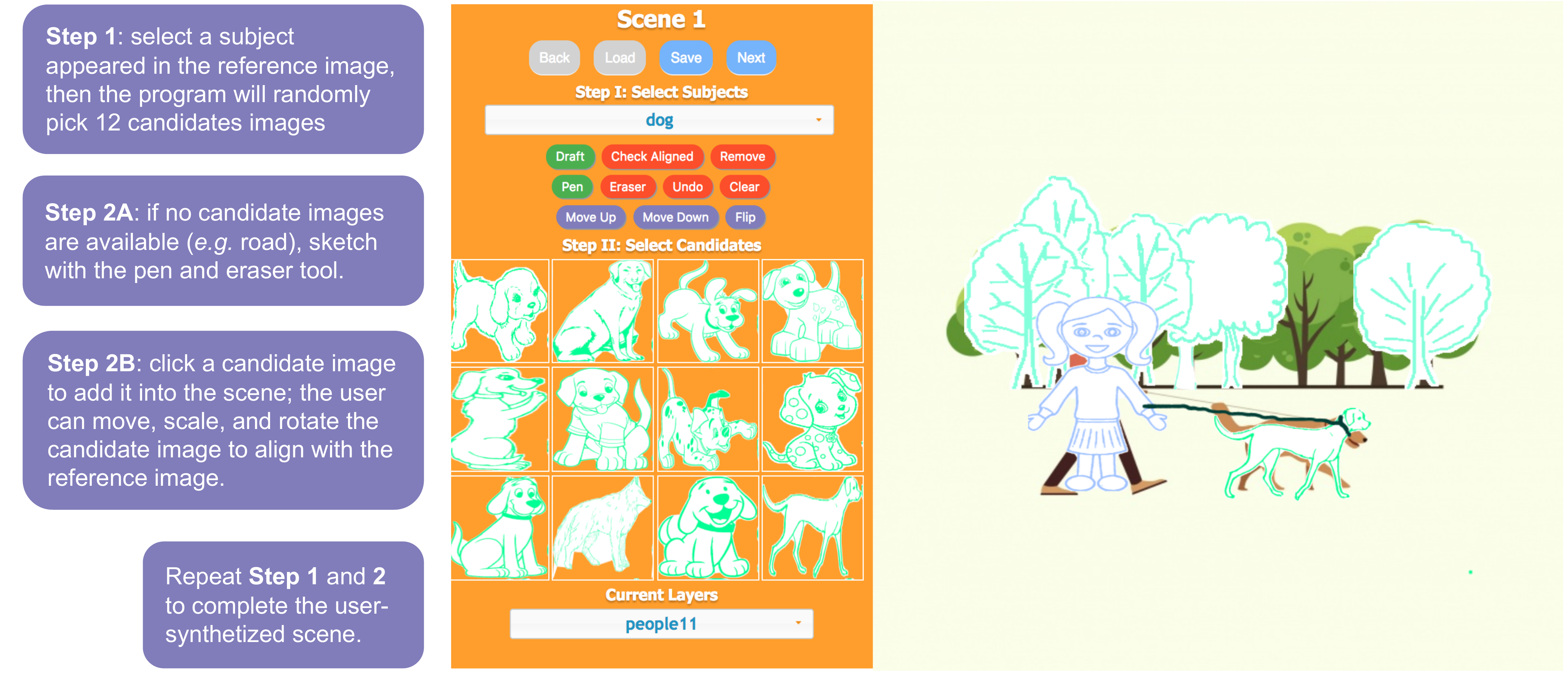}
     \caption{Interface and work flow of \textit{USketch} for crowdsourcing the dataset. See areas of function buttons (upper left), component display (lower left), and canvas (right). 
     }
     \label{fig:dataset_website}
 \end{figure}

\noindent\textbf{Step2: Scene Sketch Synthesis.}
To boost the efficiency of human creators, we devised a customary, web-based application for sketch scene synthesis. 
About 80 workers were employed to create scene sketches. Figure ~\ref{fig:dataset_website} shows the interface of the application (named ``USketch"). 

As explained before, we facilitated the creation of the sketchy scene images by allowing the worker to drag, rotate, scale, and deform the component sketch with the guidance of the reference image. The process is detailed in Fig.~\ref{fig:dataset_website}. It's worth noting that (1) we provided different sets of component sketches (even the same category) to different workers, to implicitly control the diversity of object sketches. Otherwise, workers tend to select the first several samples from the candidate pool; (2) We required the workers to produce as various occlusions as possible during the scene synthesis. This is to simulate the real scenarios and facilitate the research in segmentation. Our server recorded the transformation and semantic labels of each scene item of resulting sketchy scenes.  

At this step, we collected one scene sketch based on each reference image, using the components from the corresponding component repository. We therefore obtained 7,264 unique scene sketches. These unique scene sketches are further used as scene templates to generate more scene sketches.

\noindent\textbf{Step3: Annotation and Data Augmentation.}
The reference image is designed to help the worker to compose the scene and  enrich the layouts of the scene sketches. However, the objects in a reference image is not necessarily included in our dataset, i.e., 45 categories. In order to facilitate the future research by providing more accurate annotations, we required workers to annotate the alignment status of each object instance. 

Given there are plenty of components in our dataset, an efficient data augmentation strategy is to replace the object sketch with the rest components from the same category. Specifically, we automatically generated another 20 scene sketches for each worker-generated scene, and asked the worker to select the 4 most reasonable scenes for each scene template of Step2. Finally, we got 29K+ sketchy scene images after data augmentation.



\begin{figure}[!t]
       \centering
       \includegraphics[width=0.98\textwidth]{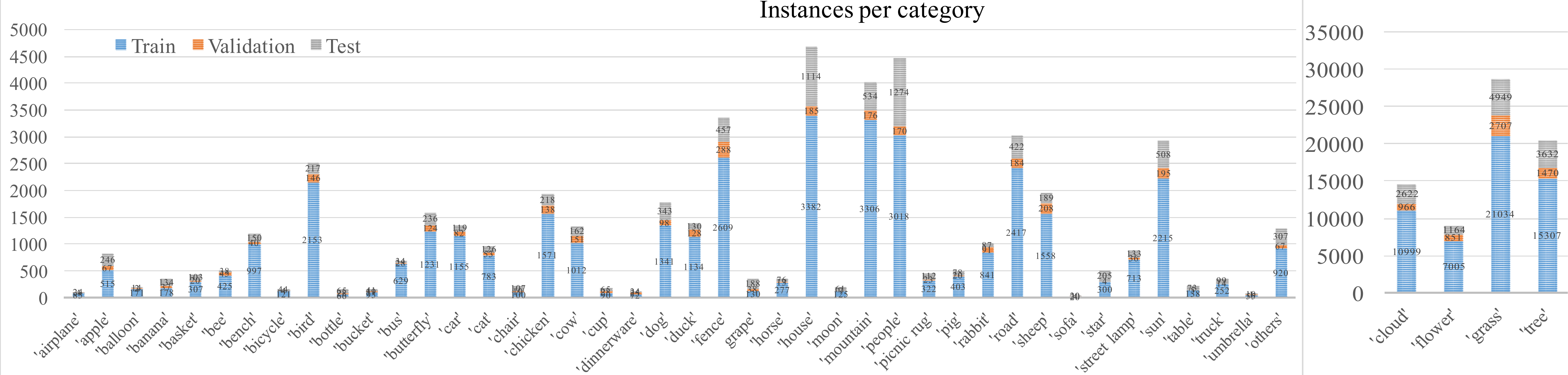}
       \caption{Object instance frequency for each category.}           
       \label{fig:statistic}
    \end{figure}

\begin{figure}[!t]
    \centering
    \includegraphics[width=0.98\textwidth]{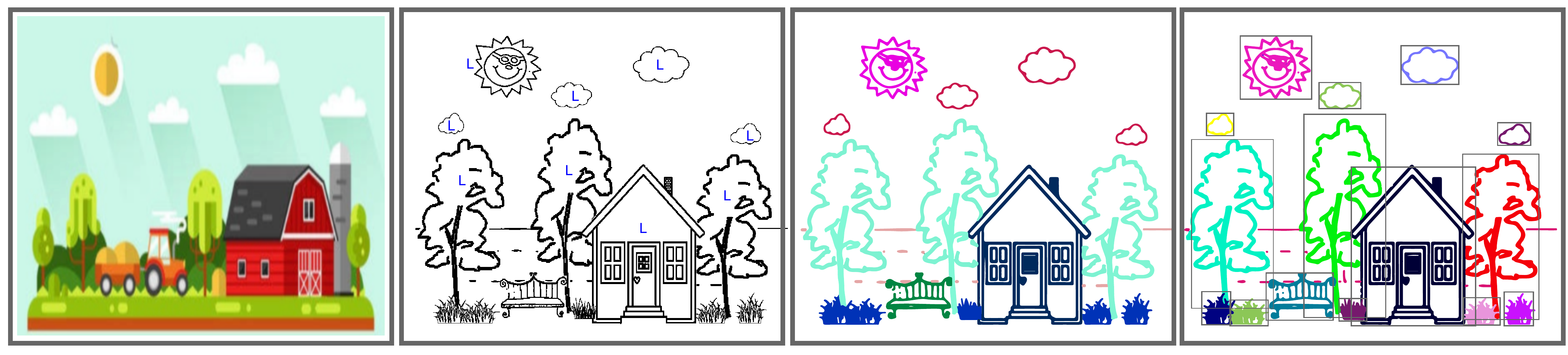}
    \caption{From left to right: reference image, synthesized sketchy scene (``L" is used to mark the category alignment), ground-truth of semantic and instance segmentation.}           
    \label{fig:labeling}
 \end{figure}

\noindent\textbf{Dataset Statistics and Analysis.}
\label{sec:dbstatistics}
To summarize, we totally obtain:
\begin{enumerate}
\item {7,264 unique scene templates created by human. Each scene template contains at least 3 object instances, where the maximum number of object instances is 94. On average there are 16 instances, 6 object classes, and 7 occluded instances per template. The maximum number of occluded instances is 66.} Figure.~\ref{fig:statistic} shows the distribution of object frequencies. 
\item 29,056 scene sketches after data augmentation (Step 3); 
\item 11,316 object sketches belonging to 44 categories. These components can be used for object-level sketch research tasks; 
\item 4730 unique reference cartoon style images which have pairwise object correspondence to the scene sketches; 
\item All sketches have 100\% accurate semantic-level and instance-level segmentation annotation (as shown in Fig.~\ref{fig:labeling}).
\end{enumerate} 

\noindent\textbf{Extensibility.}
With the scene templates and sketch components provided in the dataset, {\sc SketchyScene} can be further augmented. (1) People can segment each sketch component to get the part-level or stroke-level information; (2) The sketch components can be replaced by sketches from other resources to generate scene sketches with more varied styles.



\section{Sketch Scene Segmentation}
\label{sec:meth}


{\sc SketchyScene} can be used to study various computer vision problems. In this section, we focus on  semantic segmentation of scene sketches by modifying an existing image segmentation model. Performance is evaluated on {\sc SketchyScene} to help us identify future research directions.


\vspace{3pt}

\noindent\textbf{Problem Definition.} 
In semantic segmentation, each pixel needs to be classified into one of the candidate classes. Specifically, there is a label space $L=\left \{ l_{1},l_{2},...,l_{K} \right \}$, $K$ refers to the number of object of stuff classes. Each sketch scene image $s=\left \{ p_{1},p_{2},...,p_{N} \right \}\in \mathbb{R}^{H\times{W}}$ contains $N=W \times H$ pixels. A model trained for semantic segmentation is required to assign a label to each pixel\footnote{In this study, we consider a sketch $s$ as a bitmap image.}. So far the definition of sketch scene segmentation is identical to that of photo image segmentation. However, different from  photos, a sketch only consists of black lines (pixel intensity value equals to 0) and white background (pixel value equals to 255). Given the fact that only black pixels convey semantic information, we define the semantic segmentation in sketchy scenes as predicting a class label for each pixel whose value is 0. Taking the second image of Fig.~\ref{fig:labeling} as an example, when segmenting trees, house, sun and cloud, all black pixels on the line segments (including contours and the lines within the contours) should be classified while the rest white pixels are treated as background. 


\noindent\textbf{Challenges.}  
Segmenting a sketchy scene is challenging due to the sparsity of visual feature. First of all, a sketch scene image is dominated by white pixels. In {\sc SketchyScene}, the background ratio is 87.83\%. The rest pixels belong to $K$ foreground classes.  The classes are thus very imbalanced. Second, segmenting occluded objects becomes much harder. In photos, an object instance often contain uniform color or texture. Such cues do not exist in a sketch scene image.  

\subsection{Formulation}
We employ a state-of-the-art semantic segmentation model developed for photo images, DeepLab-v2 \cite{CP2016Deeplab}, which is customized for segmenting scene sketches. 
DeepLab-v2 has three key features, including atrous convolution, spatial pyramid spatial pooling (ASPP), and utilizing fully-connected CRF as post-processing. It is a FCN-based \cite{long2015fully} model, i.e., adapting a classification model for segmentation by replacing the final fully connected layer(s) with fully convolutional layer(s). For each input sketch, the output is a $K\times h\times w$ tensor, $K$ represents the number of class while $h\times w$ are the output segmentation  dimension. A common per-pixel softmax cross-entropy is used during training. 

Among the three features, fully-connected CRF, or denseCRF, is widely used in segmentation as a post-process. 
However, there are large blank areas in scene sketches which should be treated differently. We show that directly applying DeepLab-v2 to model the sketches results in inferior performance, and denseCRF further degrades the coarse segmentation results (see Sec.\ref{sec:background}).

Based on the characteristics of sketchy scenes, we propose to ignore the background class during modeling. This is because (1) the ratio of background pixels is much higher than the non-background pixels, which may introduce bias into the model; (2) the background information is provided in the input image and we can filter them out easily by treating the input as a mask after segmentation.  Specifically, in our implementation, the background pixels do not contribute to the loss during training. During the inference, these background pixels are assigned a non-background class label, followed by a denseCRF for refinement. Finally, the background pixels are filtered out by the input image. 

\subsection{Experiments}
\label{sec:exp}
We conducted all the experiments on {\sc SketchyScene}, using the set of 7,264 unique scene sketch templates which are split into training (5,616), validation(535), and test(1,113). Microsoft COCO is employed to verify the effectiveness of pre-training.



\noindent\textbf{Implementation details.} 
We use Tensorflow 
and ResNet101 as the base network. The initial learning rate is set to 0.0001 and mini-batch size to 1. We set the maximum training iterations as 100K and the optimiser is Adam. We keep the data as their original size (750$*$750), without applying any data augmentation on the input as we are not targeting optimal performance. We use deconvolution to scale the prediction to the same size as the ground truth mask. For denseCRF, we set the hyper parameters $\sigma_{\alpha}, \sigma_{\beta}, \sigma_{\gamma}$ to 7, 3, and 3, respectively. 

\noindent\textbf{Competitors.} 
We compare four existing models for segmenting natural photos: FCN-8s\cite{chen2017rethinking}, SegNet\cite{badrinarayanan2017segnet}, DeepLab-v2\cite{CP2016Deeplab} and DeepLab-v3\cite{long2015fully}. FCN-8s is the first deep segmentation model adapted from deep classification. It further combines coarse and fine features from different layers to boost performance. SegNet employs an encoder-decoder architecture which modifies the upsampling process to generate a more accurate segmentation result. DeepLab-v2 employs atrous convolution and denseCRF for segmentation, as explained in Sec.\ref{sec:meth}. Compared with DeepLab-v2, DeepLab-v3 incorporates global information and batch normalization, achieving comparable performance as DeepLab-v2 without employing denseCRF for refinement. In our experiment, FCN-8s and SegNet use VGG-16 while both DeepLab-v2 and v3 use ResNet101 as the base network. For fair comparison, we apply the same data processing strategy in all four models.

\noindent\textbf{Evaluation Metrics.} 
Four metrics are used to evaluate each model: Overall accuracy (OVAcc) indicates the ratio of correctly classified pixels; Mean accuracy (MeanAcc) computes the ratio of the correctly classified pixels over all classes; Mean Intersection over Union (MIoU), a commonly used metric for segmentation, computes the ratio between the intersection and union of two sets, averaged over all classes; FWIoU improves MIoU slightly by adding a class weight. 

\begin{table}[!t]
\begin{center}
\caption{Comparison of DeepLab-v2 and other baselines (\%)}
\label{table:baseline}
\begin{tabular}{c|cc|cc|cc|cc}
\hline\noalign{\smallskip}
\multirow{2}{*}{Model} & \multicolumn{2}{c|}{OVAcc} & \multicolumn{2}{c|}{MeanAcc} & \multicolumn{2}{c|}{MIoU} & \multicolumn{2}{c}{FWIoU} \\
\noalign{\smallskip}
\cline{2-9}
& val & test & val & test & val & test & val & test\\ 
\hline
\noalign{\smallskip}
FCN-8s  & 83.38 & 73.78 & 62.82 & 57.80 & 45.26 & 39.16 & 73.63 & 60.16\\
SegNet  & 84.61 & 78.61 & 58.29 & 54.05 & 42.56 & 38.32 & 76.28 & 67.91\\
DeepLab-v3  & 92.71 & 88.07 & 82.83 & \textbf{76.40} & 73.03 & \textbf{63.69} & 86.71 & 79.19\\
\hline
DeepLab-v2(final) & \textbf{92.94} & \textbf{88.38} & \textbf{84.95} & 75.92 & \textbf{73.49} & 63.10 & \textbf{87.10} & \textbf{79.76}\\
\hline
\end{tabular}
\end{center}
\end{table}

\begin{table}[!t]
\begin{center}
\caption{Comparison of including/excluding background (\%)}
\label{table:bg}
\begin{tabular}{c|cc|cc|cc|cc}
\hline\noalign{\smallskip}
\multirow{2}{*}{Model} & \multicolumn{2}{c|}{OVAcc} & \multicolumn{2}{c|}{MeanAcc} & \multicolumn{2}{c|}{MIoU} & \multicolumn{2}{c}{FWIoU} \\
\noalign{\smallskip}
\cline{2-9}
& val & test & val & test & val & test & val & test\\ 
\hline
\noalign{\smallskip}
with BG (train\&test) & \textbf{95.38} & \textbf{94.22} & 42.48 & 34.56 & 38.34 & 30.05 & \textbf{91.29} & \textbf{89.34}\\
with BG (train) & 90.21 & 86.41 & 73.54 & 66.49 & 61.50 & 52.58 & 82.67 & 77.09\\
\hline
w/o BG (final) & 92.94 & 88.38 & \textbf{84.95} & \textbf{75.92} & \textbf{73.49} & \textbf{63.10} & 87.10 & 79.76\\
\hline
\end{tabular}
\end{center}
\end{table}

\begin{table}[!t]
\begin{center}
\caption{Comparison of pre-training strategy (\%)}
\label{table:pretrain}
\begin{tabular}{c|cc|cc|cc|cc}
\hline\noalign{\smallskip}
\multirow{2}{*}{Model} & \multicolumn{2}{c|}{OVAcc} & \multicolumn{2}{c|}{MeanAcc} & \multicolumn{2}{c|}{MIoU} & \multicolumn{2}{c}{FWIoU} \\
\noalign{\smallskip}
\cline{2-9}
& val & test & val & test & val & test & val & test\\ 
\hline
\noalign{\smallskip}
Variant-1 & \textbf{93.07} & \textbf{88.67} & 82.23 & 74.97 & 71.41 & 62.12 & \textbf{87.42} & \textbf{80.19}\\
Variant-2 & 91.22 & 87.08 & 76.91 & 71.70 & 65.41 & 57.81 & 84.36& 78.01\\
Variant-3 & 91.47 & 86.44 & 79.17 & 72.24 & 67.91 & 58.54 & 84.80 & 77.18\\
\hline
Pre-ImageNet (final) & 92.94 & 88.38 & \textbf{84.95} & \textbf{75.92} & \textbf{73.49} & \textbf{63.10} & 87.10 & 79.76\\
\hline

\end{tabular}
\end{center}
\end{table}

\noindent\textbf{Comparison.}
Table~\ref{table:baseline} compares the performance of different baseline models on the new task. Clearly, both DeepLab-v2 and DeepLab-v3 perform much better than FCN and SegNet. However, DeepLab-v3 yielded similar performance as DeepLab-v2, indicating that contextual information does not have much effect for the task. This can be explained by the sparsity of sketchy scenes, and the fact that the structures in a sketch are more diverse than those in natural photos. Thus contextual information is less important than that in natural images. 

\begin{figure}[!t]
     \centering
     \includegraphics[width=0.95\textwidth]{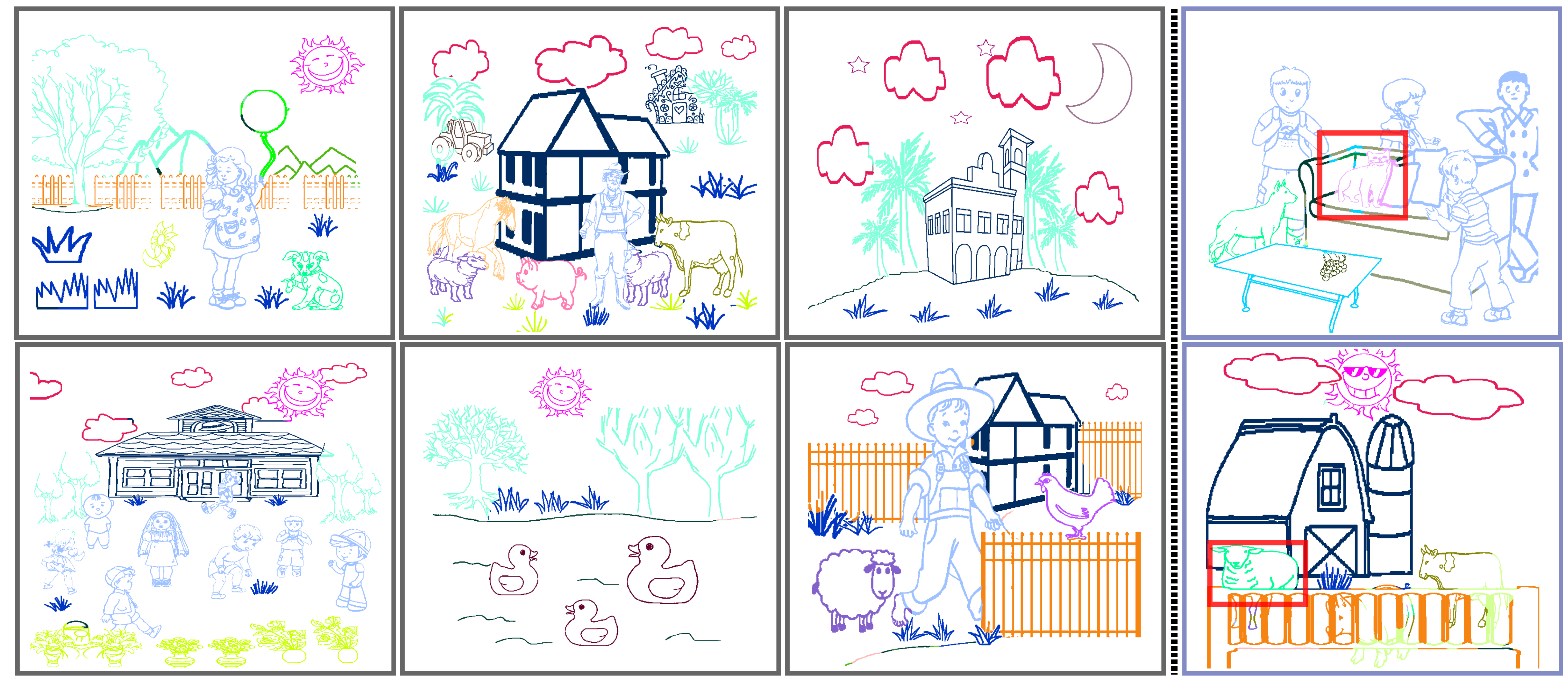}
     \caption{Visualizations of our segmentation results. Left: 6 examples with good segmentation results; right: two failure cases.}
     \label{fig:vis}
  \end{figure}

\noindent\textbf{Qualitative Results.} Figure~\ref{fig:vis} shows several segmentation results generated by DeepLab-v2 (each class is highlighted by a different colour). Although the results are encouraging, there is still  large space to improve. In particular, the failure are mainly caused by two reasons: (1) The intra-class variation is large while sketch itself is significantly deformed. For example, in the bottom image of the fourth column of Fig.~\ref{fig:vis}, the ``sheep" (highlighted in purple) is classified as a ``dog" (highlighted in green); (2) occlusions between different object instances or instances being spatially too close. As shown in the top image of the fourth column, ``cat", ``human" and ``sofa" are clustered together, making the pixels in the junction part  classified wrongly. Since sketches are sparse in visual cues and we only utilize pixel-level information, it would thus be helpful to integrate object-level information. Note that the second problem is more challenging and sketch-specific. In photo images, pixels on the contours are typically ignored. However, they are the only pixels of interest in the new task. Therefore, some sketch-specific model design need to be introduced. For examples, some perceptual grouping principles \cite{qi2015making} can be introduced to remedy this problem. See more segmentation results in the Supplementary Material.

\begin{figure}[!t]
    \centering
    \includegraphics[width=0.95\textwidth]{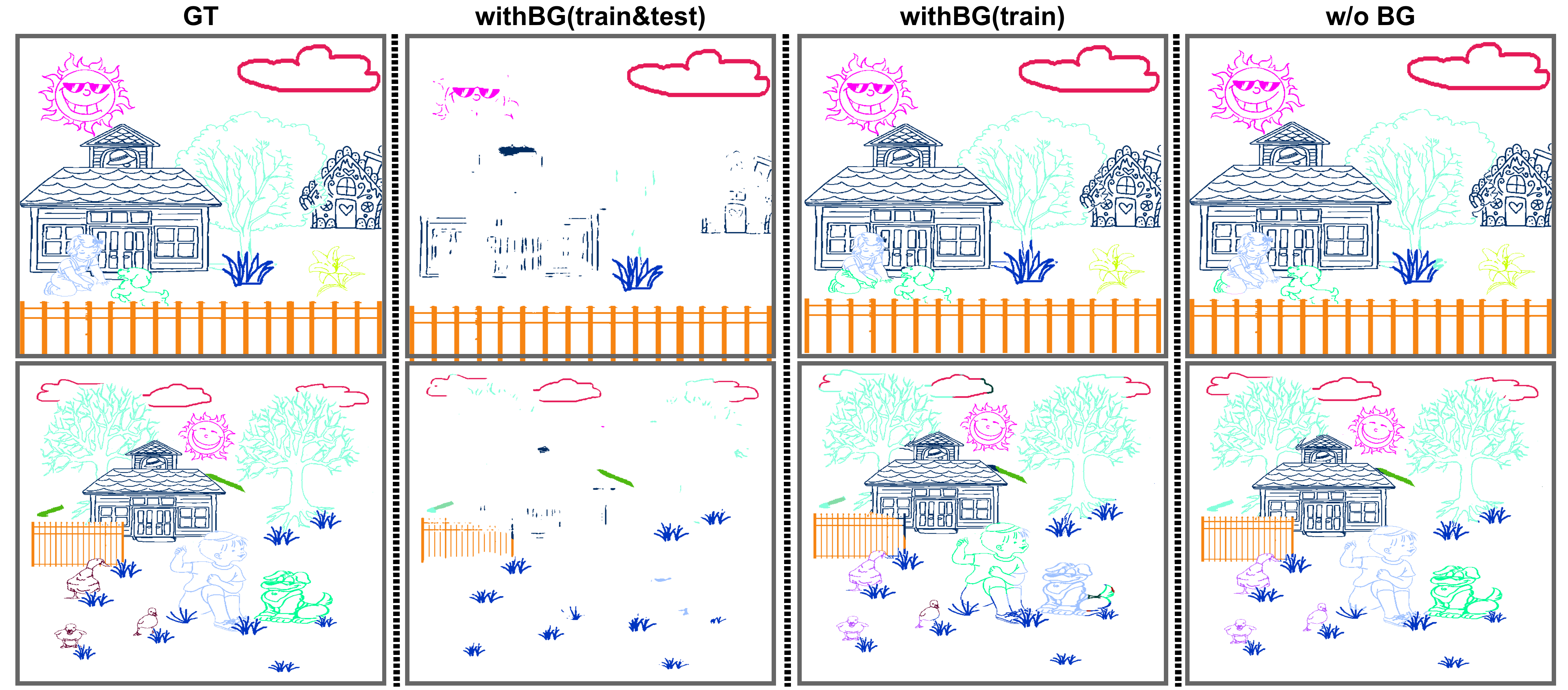}
    \caption{Comparing segmentation results when including/excluding background (BG).}         
    \label{fig:BG}
 \end{figure}

\noindent\textbf{Effect of Background.}
\label{sec:background}
As discussed earlier, the large area of background is the key problem to be solved for sketchy scene segmentation. We propose to ignore the background class during model training. When considering the background class, it mainly affects two processes, modeling by deep network and refinement by denseCRF. So we decoupled them and conducted the following experiments: (1) \textbf{withBG (train\&test)}: considering the background during training the deep model and applying denseCRF for refinement, and (2) \textbf{withBG (train)}: only consider the background during training but ignore this class for refinement, i.e., when generating the coarse segmentation, the model assigns the background class pixels a non-background class label and then feed it to denseCRF. Table~\ref{table:bg} compares the performance. We can make the following observations: (1) The performance measured in both Mean Accuracy and MIoU have a significant improvement when excluding background as a class. The Overall Accuracy and FWIoU drop since the accuracy on ``background" class is much higher than other classes; (2) The processing of background mainly affects the performance of denseCRF. This is as expected because it infers the label for each pixel by considering the neighbor pixels, thus the classes which have large ratio in the images tend to spread out. 
Some qualitative results are shown in  Fig.~\ref{fig:BG}. From the images shown in the second column,  we can see with the refinement of denseCRF, lots of pixels are merged into ``background". The last image shows the result following our proposed data processing.
 
\noindent\textbf{Effect of Pre-training.}
\label{sec:pretrain}
Our final model is pre-trained on ImageNet and fine-tuned on {\sc SketchyScene}. In this experiment, we implemented three pre-training variants: (1) \textbf{Variant-1}: Based on the ImageNet pre-trained model, we further pre-trained on the large-scale natural image segmentation dataset, i.e., Microsoft COCO, then fine-tuned on {\sc SketchyScene}. (2) \textbf{Variant-2}: Instead of pre-training on the natural images, we trained the model on edge maps extracted from the COCO dataset. In this variant, the mask of each object is region-based, i.e., with inner region pixels. (3)\textbf{Variant-3}: To simulate the target task, we further remove the inner region pixels of the mask used in Variant-2. That is, the mask covers the edges only, which is more similar to our final task. Table~\ref{table:pretrain} shows: (1) pre-training on COCO does not help. This is probably due to the large domain gap between the sketch and the natural photo. (2) Pre-training on edge maps (no matter what kind of mask they use) does not bring benefits either. Again this is due to the domain gap: different from sketches, edge maps contain lots of noise. (3) Variant-3 outperforms Variant-2, which is as expected since Variant-3 is more similar to our final task.

\section{Other Applications using {\sc SketchyScene}}
In this section, we propose several interesting applications which are enabled by our {\sc SketchyScene} dataset. 

\begin{figure}[!t]
     \centering
     \includegraphics[width=\textwidth]{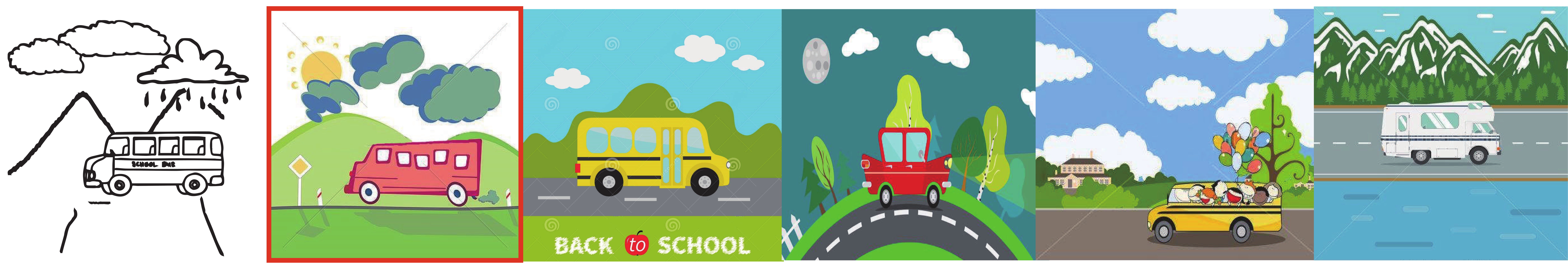}
     \caption{Retrieval results. The corresponding reference image is highlighted by red box.}
     \label{fig:retrieval}
  \end{figure}

\noindent\textbf{Image retrieval.}
Here we demonstrate an application of scene-level SBIR, which complements conventional SBIR \cite{yu2016sketch,sketchy2016,hu2013performance} by using scene-level sketches to retrieve images.
Considering the objects presented in the sketches of {\sc SketchyScene} are not 100\% aligned with the reference images (as explained in Sec.\ref{sec:datasets}), we selected sketch-photo pairs whose semantic-IoU are higher than 0.5 (2,472 pairs for training and validation while 252 for testing). Here semantic-IoU refers to the category-level overlap between the scene sketch and reference image. We develop a triplet ranking network similar to \cite{yu2016sketch} by changing the base network to InceptionV3 \cite{szegedy2016rethinking}, and adding a sigmoid cross-entropy loss as the auxiliary loss (this is to utilize the object category information to learn a more domain-invarint feature). We report accuracy at rank1 (acc.@1) and rank10 (acc.@10) inline with other SBIR papers. Overall, we obtain 32.13\% on acc.@1 and 69.48\% on acc.@10. Fig.~\ref{fig:retrieval} offers an example qualitative retrieval results.

\begin{figure}[!t]
     \centering
     \includegraphics[width=1.0\textwidth]{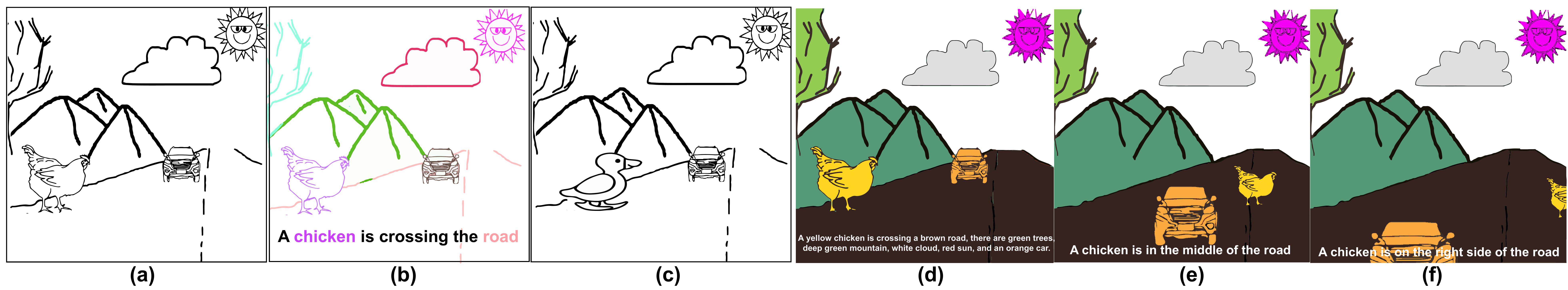}
     \caption{Applications: captioning (b), editing (c), colorization (d), and dynamic scene synthesis (d-f).}           \label{fig:app}
\end{figure}

\noindent\textbf{Sketch captioning and editing.}
%
Here we demonstrate two simple applications, namely sketch captioning and sketch editing (illustrated in Fig. \ref{fig:app}(b) and (c), respectively).  
The assumption is, based on the segmentation results, with extra annotations like image description, an image captioning model can be developed based on {\sc SketchyScene}. Furthermore, people can edit a specific object using the instance-level annotations. As shown in Fig. \ref{fig:app}(c), the ``chicken'' can be changed to a ``duck'' while the other objects are kept the same. Both of these two applications could be useful for children education.

\noindent\textbf{Sketch colorization.}
Here we show the potential of using our dataset to achieve automatic sketch colorization, when combined with the recognition and segmentation engine developed in Sec.~\ref{sec:exp}. 
In Figure~\ref{fig:app}(d), we show a colorization result by assigning different colors to different segmented objects, taking into account of their semantic labels (e.g., the sun is red).
%
%

\noindent\textbf{Dynamic scene synthesis.}
Finally, we demonstrate a more advanced application, dynamic sketch scene synthesis. We achieve this by manipulating our scene templates to construct a series of frames which are then colorized coherently across all frames. Fig.~\ref{fig:app}(d)-(f) depicts an example, ``chicken road crossing''.
%
%

\section{Conclusion, Discussion, and Future Work}
In this paper, we introduce the first large-scale dataset of scene sketches, termed {\sc SketchyScene}. It consists of a total of 29,056 scene sketches, generated using 7,264 scene templates and 11,316 object sketches. Each object in the scene is further augmented with semantic labels and instance-level masks. The dataset was collected following a modular data collection process, which makes it highly extensible and scalable. We have shown the main challenges and informative insights of adapting multiple image-based segmentation models to scene sketch data. There are several promising future directions to further enhance our scene sketch dataset, including adding scene-level annotations and text captions  to enable applications such as text-based scene generation. 

\noindent{\textbf{Acknowledgment.} {This work was partially supported by the China National 973 Program (2015CB352501), NSFC-ISF(61561146397), NSERC 611370, and the China Scholarship Council (CSC).}

\bibliographystyle{splncs04}
\bibliography{egbib}
\end{document}